# Identifying Key Features for Establishing Sustainable Agro-Tourism Centre: A Data Driven Approach

Alka Gadakh[1], Vidya Kumbhar[2*], Sonal Khosla[3], Kumar Karunendra[4]

[1] Symbiosis International (Deemed University), Pune 412115, India
[2] Symbiosis Institute of Geoinformatics, Symbiosis International (Deemed University), Pune 412115, India
[3] Odia Generative AI, Berlin 10825, Germany
[4] Symbiosis Institute of Technology, Symbiosis International (Deemed University), Pune 412115, India

Corresponding Author Email: kumbharvidya@gmail.com

**ABSTRACT**

Agro-tourism serves as a strategic economic model designed to facilitate rural development by diversifying income streams for local communities like farmers while promoting the conservation of indigenous cultural heritage and traditional agricultural practices. As a very booming subdomain of tourism, there is a need to study the strategies for the growth of Agro-tourism in detail. The current study has identified the important indicators for the growth and enhancement of agro-tourism. The study is conducted in two phases: identification of the important indicators through a comprehensive literature review and in the second phase state-of-the-art techniques were used to identify the important indicators for the growth of agro-tourism. The indicators are also called features synonymously, the machine learning models for feature selection were applied and it was observed that the Least Absolute Shrinkage and Selection Operator (LASSO) method combined with, the machine Learning Classifiers such as Logistic Regression (LR), Decision Trees (DT), Random Forest (RF) Tree, and Extreme Gradient Boosting (XGBOOST) models were used to suggest the growth of the agro-tourism. The results show that with the LASSO method, LR model gives the highest classification accuracy of 98% in 70-30% train-test data followed by RF with 95% accuracy. Similarly, in the 80-20% train-test data LR maintains the highest accuracy at 99%, while DT and XGBoost follow with 97% accuracy. The study has identified the key demographic indicators influencing the success of Agro-tourism includes family involvement, professional training, self-help group participation, adoption of self-sustainability measures, geographical area, proximity to the nearest city, transportation, availability of nearby natural resources, facilities and activities offered, and provision of regional food-demonstrating their greater importance. Environmental factors such as water resources, rainwater harvesting, electricity and energy generation, biodiversity conservation, and local agricultural yield are also found to be crucial. Economic feasibility is significantly impacted by domestic expenditure patterns.

## 1. INTRODUCTION

Tourism is a vital economic sector that contributes significantly to the global economy and sustainable development. It acts as a growth engine by stimulating investment in infrastructure, preserving cultural heritage, and fostering environmental conservation [1]. Agro-tourism has emerged as a niche sector that combines agricultural activities with tourism experiences. Agro-tourism allows visitors to engage in farming practices, learn about rural lifestyles, and enjoy farm-to-table experiences [2].

Mainstream tourism practices can be effectively integrated with Agro-tourism by leveraging its unique characteristics and aligning them with broader tourism trends. Numerous researchers have explored the potential of Agro-tourism and its ability to diversify rural economies, promote sustainable agriculture, and attract eco-conscious travelers [3-6].

It supports local farmers by generating additional income while preserving traditional farming practices [1]. It helps for sustainable rural development and also highlights economic, environmental and social benefits to rural population, especially in the regions suffering from high unemployment, rural depopulation, and limited infrastructure [5].

The researchers have explored conceptually the indicators required for the development of Agro-tourism [7-10]. These studies have only focused on potential indicators suggested for Agro-tourism, however, the data driven approach for prioritization and identification of important indicators is lacking. This indicates a strong need for exploring the growth of Agro-tourism. For the sustainable growth of Agro-tourism, it is important to identify its key indicators.

The current study aims to identify indicators for establishing and enhancing Agro-tourism.

## 2. LITERATURE REVIEW

An extensive literature survey has been done as a part of the study to find the potential parameters for Agro-tourism. Different authors have studied the role of different factors for the promotion of Agro-tourism. These have been compiled and used together for the study.

Authors identify demographic indicators such as family involvement, accommodations, nature and local attractions, traditional food, cultural events, farm activities, natural resources, livestock, and mode of transportation. Researchers also identify environmental indicators such as agriculture land and biodiversity conservation [8, 11-13].

Researcher [14] stated that the attractiveness and sustainability of Agro-tourism depend on the availability of infrastructure, involvement of the local community, socio-demographic characteristics, promotion of local food and beverages, local events, and opportunities for selling regional products. The previous study identifies demographic indicators like personal profile, services availability, accommodation, area, transportation, road accessibility, telephone coverage, shopping centers, food and beverages, and local attractions are used to evaluate potential of Agro-tourism [3]. This study also explores environmental indicators such as clean air, climate, scenic beauty, natural resources (flora, fauna, rivers), pollution levels, cultural exchange and heritage preservation as well as social benefit indicators like community involvement, local employment opportunities [3].

The study [15] aims to explore strategies for developing community-based Agro-tourism by analyzing its attractions, visitor perceptions, and internal or external conditions. The study has used demographic indicators like social aspects, community participation, local skills, cultural traditions as well as environmental factors such as agricultural diversity, natural landscapes. Sustainability plays a crucial role in shaping the attractiveness and development potential of Agro-tourism destinations [15]. The paper [16] investigates the profiles and cash flow performances of Agro-tourism businesses and examines how owners' socioeconomic characteristics affect business effectiveness. In this research, economic indicators like domestic expenses, food expenses, employee expenses, electricity expenses, and initial expenses were used to find the growth of Agro-tourism as a business enterprise.

In another study author [1] stated that sustainable practices such as eco-friendly accommodation and organic farming demonstrations can attract eco-conscious travelers while ensuring long-term viability. Preservation of local biodiversity and natural landscapes is possible through Agro-tourism activities. For sustainability of Agro-tourism, efficient use of water resources, energy, and other natural resources, including green technologies for irrigation and renewable energy solutions are equally important [17]. The study [18] suggests that demographic indicators such as natural resources, community engagement, eco-friendly accommodations, promoting local food and cultural heritage promotion are essential in sustainability of Agro-tourism. This study also states that environmental sustainability, social outreach are essential indicators for measuring Agro-tourism success.

Comprehensive literature review suggests that demographic, environmental, economical, and socio-cultural indicators are playing an important role in growth of Agro-tourism (Tables 1-3). These indicators are further used for designing machine learning models for the growth of Agro-tourism discussed in the methodology.

**Table 1.** Agro-tourism demographic features

| Abbreviation | Feature Name |
|---|---|
| D1 | Family Involvement |
| D2 | Professional Training |
| D3 | MTDC Approved |
| D4 | MART Collaboration |
| D5 | Self Help Group Partition |
| D6 | Measures Of Self-Sustainability |
| D7 | Spread over Geographical area |
| D8 | Distance from Nearest City (Km) |
| D9 | Transport facility available |
| D10 | Nearby Natural Resources |
| D11 | Facilities provided |
| D12 | Activities provided |
| D13 | Accommodation provided |
| D14 | Land Type |
| D15 | Food provided |
| D16 | Medical Service |
| D17 | Social media usage |

**Table 2.** Agro-tourism environmental features

| Abbreviation | Feature Name |
|---|---|
| EV1 | Water Sources |
| EV2 | Water Usage |
| EV3 | Rainwater Harvesting |
| EV4 | Water Conservation Methods |
| EV5 | Electricity Consumption |
| EV6 | Electricity Generation |
| EV7 | Energy Generation |
| EV8 | Atmospheric Pollution |
| EV9 | Agri-Land Conservation |
| EV10 | Biodiversity Conservation |
| EV11 | Local Agricultural Yield |

**Table 3.** Agro-tourism economical and socio-cultural features

| Category | Abbreviation | Feature Name |
|---|---|---|
| Economical | EC1 | Average/Day Expenses (Food Expenses/Tourist) |
| | EC2 | Employee Expense/Tourist |
| | EC3 | Domestic Expenses/Tourist |
| | EC4 | Electricity Consumption/Tourist |
| | EC5 | Initial Investment |
| | EC6 | Insurance Coverage |
| | EC7 | Government Subsidiary |
| Socio-cultural | SC1 | Benefits To Local Community (Employment, Market, Improvement in Socio-Economics) |
| | SC2 | Local Cultural Heritage Promotion |
| | SC3 | Safety Measure |
| | SC4 | Average no. of Tourist/Year |

## 3. METHODOLOGY OF THE STUDY

The research methodology adopted for the study is divided into two main phases. In Phase I, an extensive literature survey is done to identify the different features important for the promotion of Agro-tourism. The literature survey has been discussed in Section 2. In Phase II, state-of-the-art techniques have been applied to identify most essential Agro-tourism indicators using machine learning as shown in Figure 1.

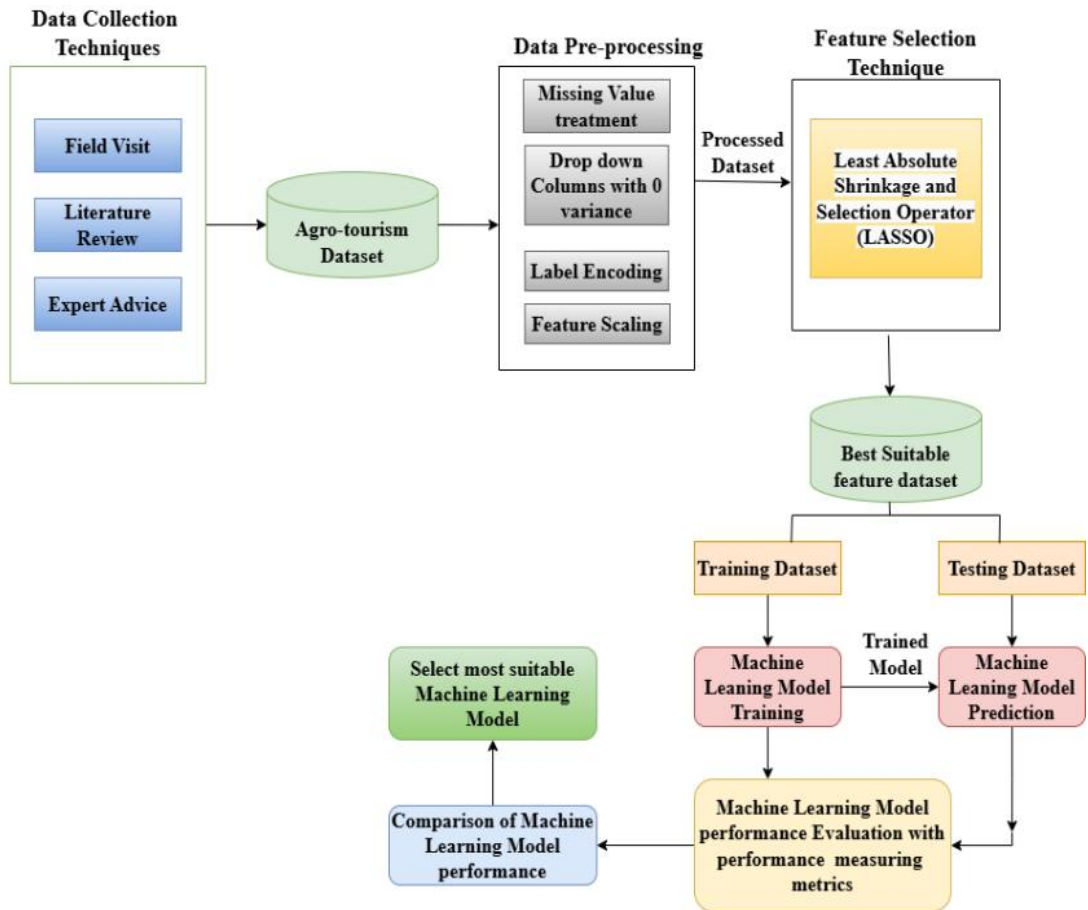

**Figure 1.** Methodology framework

### 3.1 Data collection

The area selected for the study was the Pune District of Maharashtra state in India. The study area recorded the highest concentration of Agro-tourism centers. Its proximity to urban populations, diversity in agricultural practices, and successful government and private initiatives make it an ideal region for studying the growth, challenges, and impact of Agro-tourism.

The data collection was done by conducting the field visits to 81 Agro-tourism centers out of 106 which is 76.4% of the study area through a structured questionnaire. The questionnaire mainly focused on collecting the data for the indicators identified in Section 1. Each question from the questionnaire was validated using Cronbach Alpha technique. The questions with Cronbach Alpha values greater than 0.77 were retained for further study.

### 3.2 Data preprocessing

The collected raw data underwent preprocessing to ensure it was in a usable format. Various techniques were applied, including handling missing values, label encoding, and feature scaling. Categorical data was converted into numerical format using appropriate encoding methods. Other preprocessing steps included missing value treatment, categorical data conversion, and feature scaling to enhance data quality and reliability for the study.

### 3.3 Feature selection

In this Section, the important indicators which are synonymously called as features in machine learning were identified. The ML methods such as filter approach, wrapper based, and embedded approach were applied to identify important features for growth of Agro-tourism. Feature importance is commonly evaluated using criteria such as information-theoretic measures, performance on a validation set, or accuracy on the training data, to determine a feature's effectiveness in distinguishing between classes [19]. Filter feature selection methods use statistical computation to assess the relevance of features to the target task. Only features that meet specific scoring criteria are retained. Wrapper methods naturally address multicollinearity by eliminating redundant features that fail to contribute unique information to the model. Embedded approaches include classifiers with built-in feature selection methods that perform variable selection during the training process. Unlike filter and wrapper methods, embedded approaches establish an interaction between feature selection and the learning process. Standard embedded methods for feature selection include decision tree-based algorithms such as Classification and Regression Trees (CART), C4.5, and Random Forest (RF) [20]. Although RF is inherently an embedded feature selection method, it is also frequently employed as a classification algorithm to evaluate feature subsets generated by filter and wrapper methods [19, 21, 22].

LASSO (Least Absolute Shrinkage and Selection Operator) is a regression technique that integrates variable selection with regularization, aiming to enhance model's predictive accuracy and interpretability [19]. LASSO applies a penalty to the modulus of the regression coefficients, effectively shrinking some of them to zero, which results in automatic feature selection [19]. In this study, to find the best features for growth of Agro-tourism embedded method LASSO is used.

## 3.4 Classifiers and predictors

Supervised machine learning plays an essential role in designing prediction models which could significantly enhance operational efficiency [21]. Generally, classification technique is used to predict the result in specific categories or classes which are generally predefined. These models can discover patterns and predict results by training algorithms on labelled datasets in which inputs connect with recognized outcomes [23]. This predictive potential has significance for early detection of the result which helps in taking different prevention measures before failures. In the current study, we used LASSO technique for the feature selection. To find the efficacy of feature selection techniques, machine learning classification models were applied. Considering the small nature of the dataset, simpler models like LR and DT were used to understand model behavior initially. Complex models like SVM and ANN were avoided since they can be unstable and overfit with limited data. The datasets were lacking the non-linearity, complexity in terms of high dimensionality which forced to select the ML models such as DT, LR, RF, and XGBoost over SVM and ANN to generate the result. The analysis was done by splitting the dataset into 70%-30% as well as 80%-20% training and testing dataset. The accuracy of these classifiers was tested using Accuracy, Precision, Recall, and F1-score [24].

Logistic Regression (LR) is one of the commonly adopted models in machine learning. It is typically employed to model binary outcomes [25]. LR predicts a categorical output based on input values, each assigned specific weights or coefficient values. The output variable, YY, follows a Bernoulli probability distribution, taking values between 0 and 1, ranging from $1-\pi$ to $\pi$. The threshold is typically set at the midpoint, with values exceeding this threshold classified as 1, and those below it classified as 0 [25].

The Decision Tree Classifier is one of the most powerful algorithms in machine learning. This method structures data in a tree-like format, where each branch represents a feature and its corresponding category, which is crucial for evaluating the outcomes [26]. Selecting the most relevant feature and placing it at the correct vertex is essential during tree splitting. This process leads to the generation of valuable features, resulting in highly effective outcomes [21].

The Random Forest algorithm is a robust machine learning technique based on decision trees. It creates multiple decision trees during the training process, each using a randomly selected subset of the dataset. Additionally, at each split within a tree, only a random subset of features is considered. This randomness in Random Forest helps to minimize overfitting and enhances the overall predictive performance of the model [21].

Ensemble learning integrates multiple models to achieve better predictive accuracy and XGBoost is a popular example of such an ensemble method [27, 28]. Ensemble algorithms are typically classified into three main types: bagging, boosting, and stacking [28]. XGBoost falls under the boosting category of ensemble methods [27, 28]. XGBoost is a next-generation boosting algorithm designed to build highly optimized models by identifying the best parameters through the minimization of an objective function.

## 3.5 Performance analysis methods

Performance evaluation metrics are essential for assessing model effectiveness. In this research, seven performance metrics are used: Accuracy (ACC), Sensitivity (SEN), Specificity (SPE), False Positive Rate (FPR), False Negative Rate (FNR), F1-Score, and Precision. These metrics are calculated using the formulas provided in Eqs. (1) to (7) [29].

$$ACC = \frac{TP+TN}{TP+FP+FN+TN} \times 100\% \quad (1)$$

$$SEN = \frac{TP+TN}{TP+FN} \times 100\% \quad (2)$$

$$SPE = \frac{TN}{TP+FP} \times 100\% \quad (3)$$

$$FPR = \frac{FP}{FP+TN} \times 100\% \quad (4)$$

$$FNR = \frac{FN}{TP+FN} \times 100\% \quad (5)$$

$$Precision = \frac{TP}{TP+FP} \times 100\% \quad (6)$$

$$F1 - Score = 2 \times \frac{SEN \times Precision}{SEN + Precision} \times 100\% \quad (7)$$

## 4. RESULTS

### 4.1 Feature selection

In Recursive Feature Elimination method XGBoost shows strong performance with highest accuracy of 69% with Receiver Operating Characteristic Curve (ROC) and Area Under the Curve (AUC) score 81% and f1 score 1 as well as 64% accuracy with 46% ROC-AUC score and recall value 1 in 70%-30% and 80%-20% data split respectively (Table 4).

**Table 4.** Recursive feature elimination method result

| Training Dataset 70% and Test Dataset 30% | | | | | |
|---|---|---|---|---|---|
| Algorithms | Accuracy | Precision | Recall | F1 score | ROC-AUC |
| LR | 0.63 | 0.82 | 0.69 | 0.75 | 0.42 |
| DT | 0.63 | 0.82 | 0.69 | 0.75 | 0.42 |
| RF | 0.63 | 0.82 | 0.69 | 0.75 | 0.42 |
| XGBoost | 0.69 | 0.44 | 0.69 | 1 | 0.81 |
| Training Dataset 80% and Test Dataset 20% | | | | | |
| Algorithms | Accuracy | Precision | Recall | F1 score | ROC-AUC |
| LR | 0.64 | 0.86 | 0.67 | 0.75 | 0.34 |
| DT | 0.64 | 0.86 | 0.67 | 0.75 | 0.34 |
| RF | 0.45 | 0.71 | 0.56 | 0.63 | 0.34 |
| XGBoost | 0.64 | 0.64 | 1 | 0.78 | 0.46 |

In Fisher's score method Random Forest gives highest accuracy of 69% with ROC-AUC score 55% as well as 73% accuracy with 45% ROC-AUC score in 70%-30% and 80%-20% data split, respectively (Table 5).

In the Information Gain method XGBoost gives highest accuracy of 62% with ROC-AUC score of 45% in 70%-30% data split. Where as in 80%-20% data split Logistic Regression, Random Forest, and XGBoost gave highest accuracy of 64% with 52%, 52%, and 50% ROC-AUC score respectively (Table 6).

In the LASSO method, Logistic Regression shows strong performance with the highest accuracy of 75% with ROC-AUC score 58% as well as 73% accuracy with 54% ROC-

2798

AUC score in 70%-30% and 80%-20% data split, respectively (Table 7). In LASSO method Random Forest also gives the highest accuracy of 73% but the precision value is 1 (Table 7). After comparing all four feature selection methods, LASSO gives the consistent results with the highest accuracy as compared to remaining methods.

**Table 5.** Fisher's score method result

| | Training Dataset 70% and Test Dataset 30% | | | | |
|---|---|---|---|---|---|
| Algorithms | Accuracy | Precision | Recall | F1 score | ROC-AUC |
| LR | 0.63 | 0.82 | 0.69 | 0.75 | 0.42 |
| DT | 0.63 | 0.82 | 0.69 | 0.75 | 0.42 |
| RF | 0.63 | 0.82 | 0.69 | 0.75 | 0.42 |
| XGBoost | 0.69 | 0.44 | 0.69 | 1 | 0.81 |
| | Training Dataset 80% and Test Dataset 20% | | | | |
| Algorithms | Accuracy | Precision | Recall | F1 score | ROC-AUC |
| LR | 0.64 | 0.86 | 0.67 | 0.75 | 0.34 |
| DT | 0.64 | 0.86 | 0.67 | 0.75 | 0.34 |
| RF | 0.45 | 0.71 | 0.56 | 0.63 | 0.34 |
| XGBoost | 0.64 | 0.64 | 1 | 0.78 | 0.46 |

**Table 6.** Information gain method result

| | Training Dataset 70% and Test Dataset 30% | | | | |
|---|---|---|---|---|---|
| Algorithms | Accuracy | Precision | Recall | F1 score | ROC-AUC |
| LR | 0.50 | 0.64 | 0.64 | 0.64 | 0.37 |
| DT | 0.50 | 0.73 | 0.62 | 0.67 | 0.37 |
| RF | 0.56 | 0.82 | 0.64 | 0.72 | 0.37 |
| XGBoost | 0.62 | 0.67 | 0.91 | 0.77 | 0.45 |
| | Training Dataset 80% and Test Dataset 20% | | | | |
| Algorithms | Accuracy | Precision | Recall | F1 score | ROC-AUC |
| LR | 0.64 | 0.86 | 0.67 | 0.75 | 0.52 |
| DT | 0.55 | 0.86 | 0.60 | 0.71 | 0.52 |
| RF | 0.64 | 0.86 | 0.67 | 0.75 | 0.52 |
| XGBoost | 0.64 | 0.67 | 0.86 | 0.75 | 0.5 |

**Table 7.** LASSO method result

| | Training Dataset 70% and Test Dataset 30% | | | | |
|---|---|---|---|---|---|
| Algorithms | Accuracy | Precision | Recall | F1 score | ROC-AUC |
| LR | 0.75 | 0.91 | 0.77 | 0.83 | 0.58 |
| DT | 0.75 | 0.91 | 0.77 | 0.83 | 0.58 |
| RF | 0.69 | 1.00 | 0.69 | 0.81 | 0.58 |
| XGBoost | 0.69 | 0.71 | 0.91 | 0.80 | 0.47 |
| | Training Dataset 80% and Test Dataset 20% | | | | |
| Algorithms | Accuracy | Precision | Recall | F1 score | ROC-AUC |
| LR | 0.73 | 0.86 | 0.75 | 0.80 | 0.54 |
| DT | 0.55 | 0.71 | 0.63 | 0.67 | 0.54 |
| RF | 0.73 | 1.00 | 0.70 | 0.82 | 0.54 |
| XGBoost | 0.64 | 0.64 | 1 | 0.78 | 0.43 |

LASSO facilitates automatic feature selection by adding a L1 penalty term to the loss function. This penalty is controlled by a regularization parameter, commonly denoted as λ in mathematical formulations, or alpha (α) in many machine learning libraries. Both determine the extent of regularization applied. As λ (or α equivalently) increases, the strength of the penalty on the regression coefficients becomes stronger, encouraging more coefficients to shrink towards zero. Coefficients that shrink to exactly zero effectively remove the corresponding features from the model to reduced model complexity, and improved interpretability.

In LASSO method, to select the best features, it is necessary to tune the α hyperparameter to achieve the optimal LASSO regression. LASSO with cross-validation is used to identify the best α value for optimal selection. The cross-validation process yields an optimal α value of 0.05 as shown in Figure 2. The LASSO technique automatically identifies and selects the relevant features, discarding the irrelevant ones by setting their coefficients to 0. Using this optimal α value, LASSO selected the top 35 most important features from the 86 independent features.

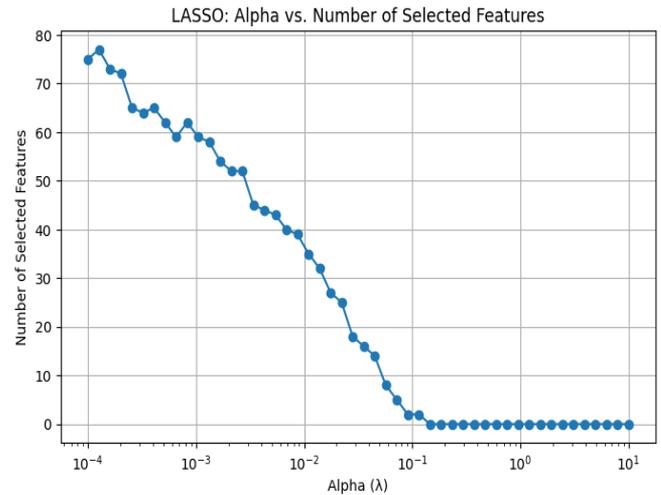

**Figure 2.** Optimal alpha with cross validation

The LASSO technique identified fourteen demographic indicators, six environmental, one economical, and one socio-cultural indicator. Demographic indicator includes family member involvement, professional training, participation in self-help groups, self-sustainability measures, the area of the Agro-tourism center, distance from the nearest city, transportation, nearby natural resources, activities, facilities, accommodation, local food, first-aid availability, and the use of social media for marketing. Environmental indicators encompass water sources, land preservation, local crop yield, and water conservation methods. Economic indicators include daily employee expenses, whereas in socio-cultural, safety measure indicator is important. These indicators are further given to ML models for finding prediction of growth of agro-tourism.

### 4.2 Prediction model results

After identification of best suitable features, the next step was to fit these features to different ML-based classifiers and find model accuracy and robustness.

The performance of these classifiers, both before and after dataset balancing, is presented in Table 8 and Table 9 for the 70-30% and 80-20% dataset splits, respectively. Initially, it was observed that models like XGBoost and RF in 70-30% and 80-20% dataset splits were performing poor due to overfitting with the accuracy of 100% (Table 8 and Table 9). It was observed that the dataset was imbalanced these classifiers were facing the issues of overfitting and for resolving this issue, data augmentation technique like Synthetic Minority Over-sampling Technique (SMOTE) was used [30]. SMOTE balances the minority sample by adding synthetic samples to the dataset. After data augmentation the data set was balanced and was applied to LR, DT, RF, and XGBoost classifiers. In binary classification models, it is common to use confusion matrix which represent prediction values by algorithm versus actual values. Figure 3 shows a confusion matrix for the selected classifiers with 70%-30% dataset split.

As presented in Table 8, before applying SMOTE balancing, the XGBoost classifier achieves the highest accuracy of 100%. However, the ROC-AUC score is undefined, indicating potential overfitting and poor generalization performance of the classifier. To address the issues of overfitting, the imbalanced dataset was balanced using SMOTE. The result indicates that after balancing, the LR classifier achieved the highest accuracy of 98% and a ROC-AUC score of 99%, demonstrating improved model performance and generalization. To check whether the performance after data augmentation was overfitted further cross validation techniques were used. By evaluating variations in key metrics across different folds, cross-validation helped to determine whether the model's high performance is due to genuine learning or excessive memorization of training data. After 5-fold-cross validation for LR we got a high mean AUC (0.98) with standard deviation score 0.03 which is less than 0.05 which suggests that model is stable. For RF, mean AUC is 0.99 and std. dev is 0.019 which confirms that the model is stable and the result is accurate. There is no condition of overfitting.

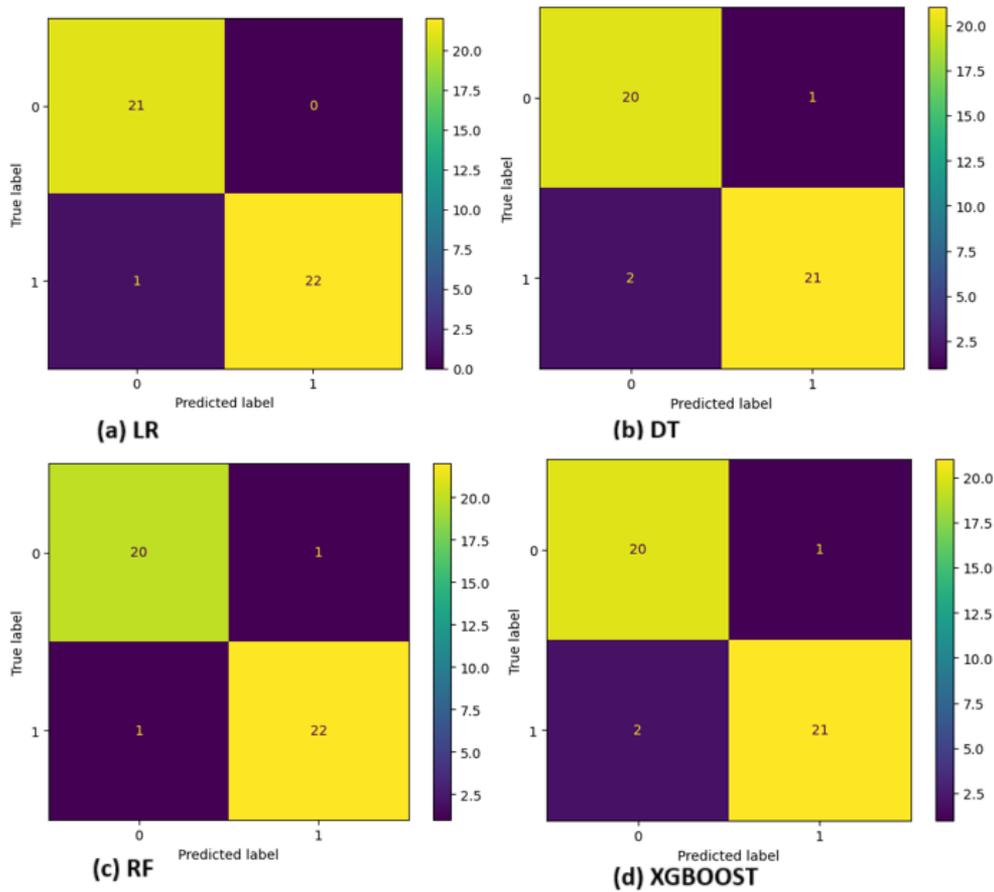

**Figure 3.** Confusion matrix for classifiers with 70%-30% dataset split

**Table 8.** Classifier analysis results with 70%-30% dataset split

| Training Dataset 70% and Test Dataset 30% | | | | | |
|---|---|---|---|---|---|
| Before Balancing Dataset | | | | | |
| Algorithm | Accuracy | Precision | Recall | F1 score | ROC-AUC |
| LR | 0.88 | 0.88 | 1 | 0.93 | NaN |
| DT | 0.84 | 0.84 | 1 | 0.91 | NaN |
| RF | 0.96 | 0.96 | 1 | 0.98 | NaN |
| XGBoost | 1 | 1 | 1 | 1 | NaN |
| After Balancing Dataset Using SMOTE | | | | | |
| Algorithm | Accuracy | Precision | Recall | F1 score | ROC-AUC |
| LR | 0.98 | 0.96 | 1.00 | 0.98 | 0.99 |
| DT | 0.93 | 0.91 | 0.95 | 0.93 | 0.95 |
| RF | 0.95 | 1.00 | 0.92 | 0.96 | 0.99 |
| XGBoost | 0.93 | 0.92 | 0.96 | 0.94 | 0.98 |

**Table 9.** Classifier analysis results with 80%-20% dataset split

| Training Dataset 80% and Test Dataset 20% | | | | | |
|---|---|---|---|---|---|
| Before Balancing Dataset | | | | | |
| Algorithm | Accuracy | Precision | Recall | F1 score | ROC-AUC |
| LR | 0.88 | 0.88 | 1.00 | 0.94 | NaN |
| DT | 0.82 | 0.82 | 1.00 | 0.90 | NaN |
| RF | 1.00 | 1.00 | 1.00 | 1.00 | NaN |
| XGBoost | 1.00 | 1.00 | 1.00 | 1.00 | NaN |
| After Balancing Dataset Using SMOTE | | | | | |
| Algorithm | Accuracy | Precision | Recall | F1 score | ROC-AUC |
| LR | 0.99 | 0.99 | 0.99 | 0.99 | 0.99 |
| DT | 0.93 | 0.94 | 0.94 | 0.94 | 0.93 |
| RF | 0.97 | 1.00 | 0.94 | 0.97 | 0.99 |
| XGBoost | 0.97 | 0.94 | 1.00 | 0.97 | 0.99 |

To evaluate reliability of model performance, we computed 95% Confidence Intervals (CI) and conducted paired t-tests on 5-fold cross-validation accuracy scores.LR demonstrated the highest accuracy with 95% CI of [0.974, 0.986], followed by RF [0.944, 0.956]. The p-values from paired t-tests revealed statistically significant differences in performance between LR and RF (p=0.000), RF and XGBoost (p < $10^{-59}$), and LR and XGBoost (p<$10^{-60}$). These findings confirm that the observed performance is improved due to SMOTE based balancing as shown in Table 8.

As shown in Table 9, before balancing the dataset, the XGBoost and RF gives highest accuracy (100%) with ROC-AUC undefined. As the recall value of all models is 1, there is a clear sign of overfitting. To validate the reliability of model comparison, statistical testing using 5-fold cross-validation with 95% CI were computed for each model.LR achieved the highest mean accuracy (0.99) with a narrow CI [0.984,0.996], indicating high stability. Paired t-tests revealed that LR significantly outperformed both RF (p=0.0032) and XGBoost (p=1.000). These results confirm that LR consistently delivers superior and statistically reliable accuracy on the balanced dataset (Table 9). After the application of SMOTE the LR gives highest accuracy of 99% with highest ROC-AUC score of 99%. Area Under Curve (AUC) is a measure used to compare these classifiers and the result is shown in Figure 4.

Further, 5-fold cross validation means accuracy of 97% with standard deviation score 0.03 states that the LR model was stable and gives the best result among all models discussed above.

The result confirms that the Logistic regression model is the most suitable model for finding the potential of Agro-tourism (Figure 5).

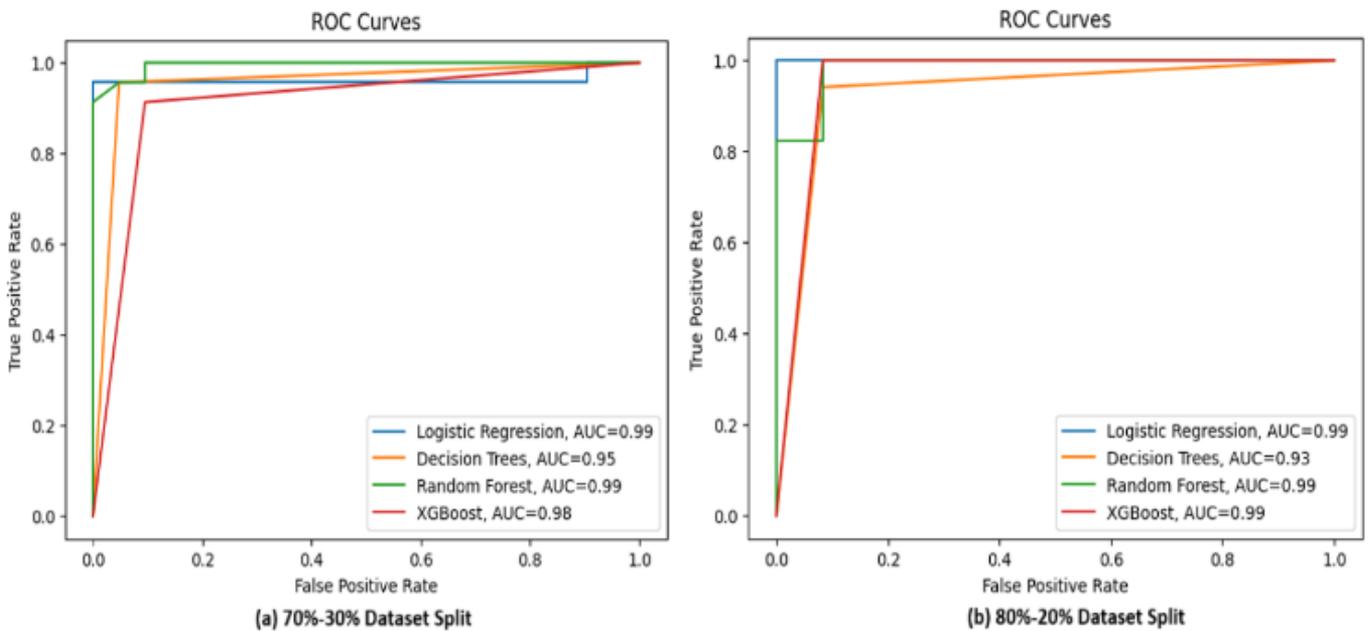

**Figure 4.** ROC-AUC results with (a) 70%-30% dataset split (b) 80%-20% dataset split

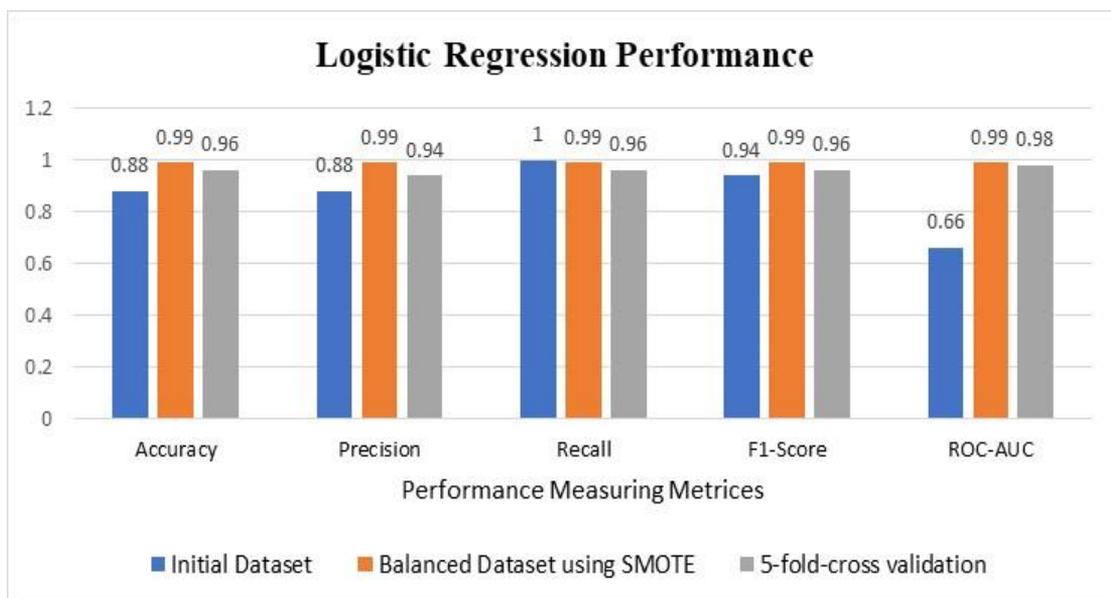

**Figure 5.** Logistic regression classifier performance

## 5. CONCLUSION

This study explores the potential growth and enhancement of Agro-tourism. The comprehensive literature review concludes that there was a limited amount of work done for the identification of important indicators, especially for Agro-tourism using state-of-the-art techniques. The current study is one of the first studies for identifying indicators and designing the machine learning model for the growth of the Agro-tourism. The LASSO method used for feature selection combined with Logistics Regression and data augmentation helped to achieve optimal results. The study concludes that the data-driven approach can effectively help to design the models for the growth of Agro-tourism. Sustainability can be effectively achieved through the integration of data-driven approach.

One of the key challenges in this research was the limited exploration of Agro-tourism using state-of-the-art techniques, as only few studies have addressed this domain with advanced analytical methods. Additionally, this study integrates perspectives from both the tourism and technical fields, offering a multidisciplinary approach that bridges the gap between these areas.

This study is limited to our study area which is Pune district from Maharashtra state of India, which may constrain the generalizability of the findings due to variability in geography, climate, and the agriculture pattern of the region. The future scope of this study includes validation of the model with more diverse datasets, and to develop an integrated agro climatic region wise model for sustainable Agro-tourism.